\documentclass[10pt,twocolumn,letterpaper]{article}

\usepackage{iccv}
\usepackage{times}
\usepackage{epsfig}
\usepackage{graphicx}
\usepackage{amsmath}
\usepackage{amssymb}
\usepackage{multirow}
\usepackage{subcaption}
\usepackage{comment}
\usepackage{authblk}

\usepackage[pagebackref=true,breaklinks=true,letterpaper=true,colorlinks,bookmarks=false]{hyperref}

\usepackage{color}

\iccvfinalcopy
\ificcvfinal\fi

\begin{document}
\title{Camera Relocalization by Computing Pairwise Relative Poses \\Using Convolutional Neural Network}



\newcommand*\samethanks[1][\value{footnote}]{\footnotemark[#1]}
\author[1]{Zakaria Laskar \thanks{Equal contribution}}
\author[1]{Iaroslav Melekhov \samethanks}
\author[2]{Surya Kalia}
\author[1]{Juho Kannala}
\affil[1]{Aalto University, Finland \authorcr
       {\tt\small firstname.lastname@aalto.fi}}
\affil[2]{Indian Institute of Technology, Delhi, India \authorcr
       {\tt\small cs5150296@iitd.ac.in}}
\renewcommand\Authands{ and }

\maketitle

\begin{abstract}
We propose a new deep learning based approach for camera relocalization. Our approach localizes a given query image by using a convolutional neural network (CNN) for first retrieving similar database images and then predicting the relative pose between the query and the database images, whose poses are known. The camera location for the query image is obtained via triangulation from two relative translation estimates using a RANSAC based approach. Each relative pose estimate provides a hypothesis for the camera orientation and they are fused in a second RANSAC scheme. The neural network is trained for relative pose estimation in an end-to-end manner using training image pairs. In contrast to previous work, our approach does not require scene-specific training of the network, which improves scalability, and it can also be applied to scenes which are not available during the training of the network. As another main contribution, we release a challenging indoor localisation dataset covering 5 different scenes registered to a common coordinate frame. We evaluate our approach using both our own dataset and the standard 7 Scenes benchmark. The results show that the proposed approach generalizes well to previously unseen scenes and compares favourably to other recent CNN-based methods.

\end{abstract}

\section{Introduction}

Camera relocalization, or image-based localization is a fundamental problem in robotics and computer vision. It refers to the process of determining camera pose from the visual scene representation and it is essential for many applications such as navigation of autonomous vehicles, structure from motion (SfM), augmented reality (AR) and simultaneous localization and mapping (SLAM). Due to importance of these problems various relocalization approaches have been proposed. Point-based localization approaches find correspondences between local features extracted from an image by applying image descriptors (SIFT, ORB, etc ~\cite{SURF,SIFT,ORB}) and 3D point clouds of the scene obtained from SfM. In turn, such set of 2D-3D matches allows to recover the full 6-DoF (location and orientation) camera pose. However, this low-level process of finding matches does not work robustly and accurately in all scenarios, such as textureless scenes, large changes in illumination, occlusions and repetitive structures. 

Recently, various machine learning methods \cite{DSAC,scorfShotton,scorfValentin}, such as scene coordinate regression forest (SCoRF)~\cite{scorfShotton,scorfValentin}, have been successfully applied to camera localization problem. SCoRF utilize predicted 3D location of four pixels of an input image to generate an initial set of camera pose hypotheses which are subsequently refined by a RANSAC loop. 
However, all these methods require depth maps associated with input images at training time, thus the applicability of such approaches is restricted. 

Inspired by the success in image classification~\cite{ResNet,ImagenetKrizh}, semantic segmentation~\cite{SegmentationHong,SegmentationNoh} and image retrieval~\cite{ImageRetrievalBabenko,retrievalGordo}, convolutional neural networks (CNNs) have also been used to predict camera pose from visual data  \cite{Posenet2,Posenet}. They cast camera relocalization as a regression problem, where camera location is directly estimated by CNN pre-trained on image classification data~\cite{ImageNet}. Although learning-based approaches overcome many disadvantages of point-based methods, they still have certain limitations. Directly regressing the absolute camera pose constrains the current machine learning models to be trained and evaluated scene-wise when the scenes are registered to different coordinate frames. 
 The reason for this is that the trained model learns a mapping from image (pixels) to pose which is dependent on the coordinate frame of the training data belonging to a particular scene. This causes complications, especially if one is interested in localization across several scenes simultaneously, and also prevents transferring learnt knowledge of geometric relations between scenes. The second problem is the obviously limited scalability to large environments since a finite neural network has an upper bound on the physical area that it can learn, as pointed out by \cite{Posenet}. 


In this paper, we propose to decouple the learning process from the coordinate frame of the scene. That is, instead of directly regressing absolute pose like \cite{Posenet2,Posenet,HourglassPose}, we train a Siamese CNN architecture to regress the relative pose between a pair of input images and propose a pipeline for computing the absolute pose from several relative pose estimates between the query image and  database images. This approach is flexible and has several benefits: \emph{(a)} our CNN can learn from image pairs of any scene thereby being able to improve towards generic relative pose estimator; \emph{(b)} a single network can be trained and used for localization in several disjoint scenes simultaneously, even in scenes whose training images are scarce or not available during the training time of the network; and \emph{(c)} the approach is scalable because a single CNN can be used for various scenes and the full scene-specific database (\ie training) images are not needed in memory at test time either as compact feature descriptors and fast large-scale image retrieval techniques can be utilized instead.


To summarize, we make the following contributions:
\begin{itemize}
\item 
 We propose a new deep learning based approach for camera relocalization. Our approach is general and scalable alternative to previous models and 
compares favourably to other CNN-based methods.
\item We show through extensive evaluation the generalization capability of the approach to localize cameras in scenes unseen during training of the CNN. 
\item We introduce a new challenging indoor dataset with accurate ground truth pose information and evaluate the proposed method also on this data.
\end{itemize}

The rest of the paper is organized as follows. Section~\ref{sec:related_work} discusses related work in image-based localization. Section~\ref{sec:pipeline} describes the network structure and the whole pipeline of our approach. The details of a new large indoor dataset and evaluation results of our method are provided in Section~\ref{sec:dataset} and Section~\ref{sec:experiments} accordingly. Conclusion and some suggestions for future work are given in Section \ref{sec:conclusion}.

We will make the source code and the dataset publicly available upon publication of the paper.

\section{Related work}\label{sec:related_work}

Camera relocalization approaches largely belong to two classes: visual place recognition methods and 3D model-based localization algorithms. Visual place recognition methods cast image-based localization problem as an image retrieval task and apply standard techniques such as image descriptors (SIFT, ORB, SURF~\cite{SURF,SIFT,ORB}), fast spatial matching~\cite{Oxford5k}, bag of visual words~\cite{BOWCula, BOWVarma} to find a representation of an unknown scene (a query image) in a database of geo-tagged images. Then, the geo-tag of the most relevant retrieved database image is considered as an approximation of a query location. The major limitation of visual recognition methods is that the images in database are often sparse, so that in situations where the query is far from database images the estimate would be inaccurate~\cite{scorfValentin}.

Structure-based localization methods utilize a 3D scene representation obtained from SfM and find correspondences between 3D points and local features extracted from a query image establishing a set of 2D-3D matches. Finally, the camera pose is established by applying RANSAC loop in combination with a Perspective-n-Point algorithm~\cite{PnPBujnak}. However, descriptor matching is expensive and time-consuming procedure making camera relocalization complicated problem for large scale scenes such as urban environment. In order to accelerate this stage,~\cite{PrioritizedSearchSnavely,PrioritizedSearchSattler} eliminate correspondence search as soon as enough matches have been found, and~\cite{SattlerTopDbSearch2,SattlerTopDbSearch1} propose to perform matching with the 3D points of top-retrieved database images.

Sattler~\etal~\cite{sattler_CVPR2017} demonstrate that combining visual place recognition approaches with local SfM leads to better localization performance compared with 3D-based methods. However, the localization process itself is still very time-consuming.

It has also been shown that machine learning methods have potential for providing efficient solutions to the pose estimation problem. Similar to structure-based localization approaches, Shotton~\etal~\cite{scorfShotton} utilize a regression forest to predict a 3D point location for each pixel of an input RGB-D image. Thus, the method establishes 2D-3D matches which are then used to recover 6-DoF camera pose by applying RANSAC. Rather then finding point correspondences, Valentin~\etal\cite{scorfValentin} propose to exploit the uncertainty of the predicted 3D point locations during pose estimation. Brachmann~\etal~\cite{DSAC} propose a differentiable RANSAC method for camera localisation from an RGB image. However, it still requires dense depth maps in the training stage. 

Recently proposed CNNs-based approaches have shown great success in image-based localization. Originally, utilizing CNNs to directly regress camera relocalization was proposed by Kendall~\etal~\cite{Posenet}. Their method, named PoseNet, adapts GoogLeNet~\cite{GoogLeNet} architecture pre-trained on large-scale image classification data to reconstruct 6-DoF camera pose from an RGB image. In the recent paper~\cite{Posenet2}, Kendall~\etal propose a novel loss function based on scene reprojection error and show its efficiency in appearance-based localization. Contrary to~\cite{Posenet2,Posenet},  HourglassPose~\cite{HourglassPose} utilizes a symmetric encoder-decoder network structure with skip connections which leads to improvement in the localization accuracy outperforming PoseNet. 
Partly motivated by advances of Recurrent Neural Networks (RNN) in text classification, Clark~\etal and Walch~\etal apply LSTM networks to determine the location from which a photo was taken~\cite{VidLoc,LSTMPose}. Following \cite{Posenet}, both of these methods, called VidLoc \cite{VidLoc} and LSTMPose \cite{LSTMPose}, utilize GoogLeNet to extract features from input images to be localized, then feeding these features to a block of LSTM units. The regression part consisting of fully-connected layers utilizes output of LSTM units to predict camera pose. The major drawback of VidLoc is that it requires a sequence of adjacent image frames as input and is able to estimate camera translation only.

The proposed approach is related to all previously discussed CNN-based methods, but it is the first one solving image-based localization problem via relative camera pose. Inspired by~\cite{MelekhovRelative,OliveiraRelative,Demon}, we apply Siamese neural network to predict relative orientation and relative translation between two views. These relative translation estimates are then triangulated to recover the absolute camera location. Compared with~\cite{VidLoc,Posenet2,Posenet,HourglassPose,LSTMPose}, our study provides more general and scalable solution to image-based localization task. That is, the proposed approach is able to estimate 6-DoF camera pose for scenes registered to different coordinates frames, unlike existing CNN-based methods. Finally, compared with traditional machine learning approaches, our method does not require depth maps for training, thus it is applicable for outdoor scenes as well. Further details of our approach will be given in the following sections.

\begin{figure*}[t!]
		\centering
        \includegraphics[width=0.9\textwidth]{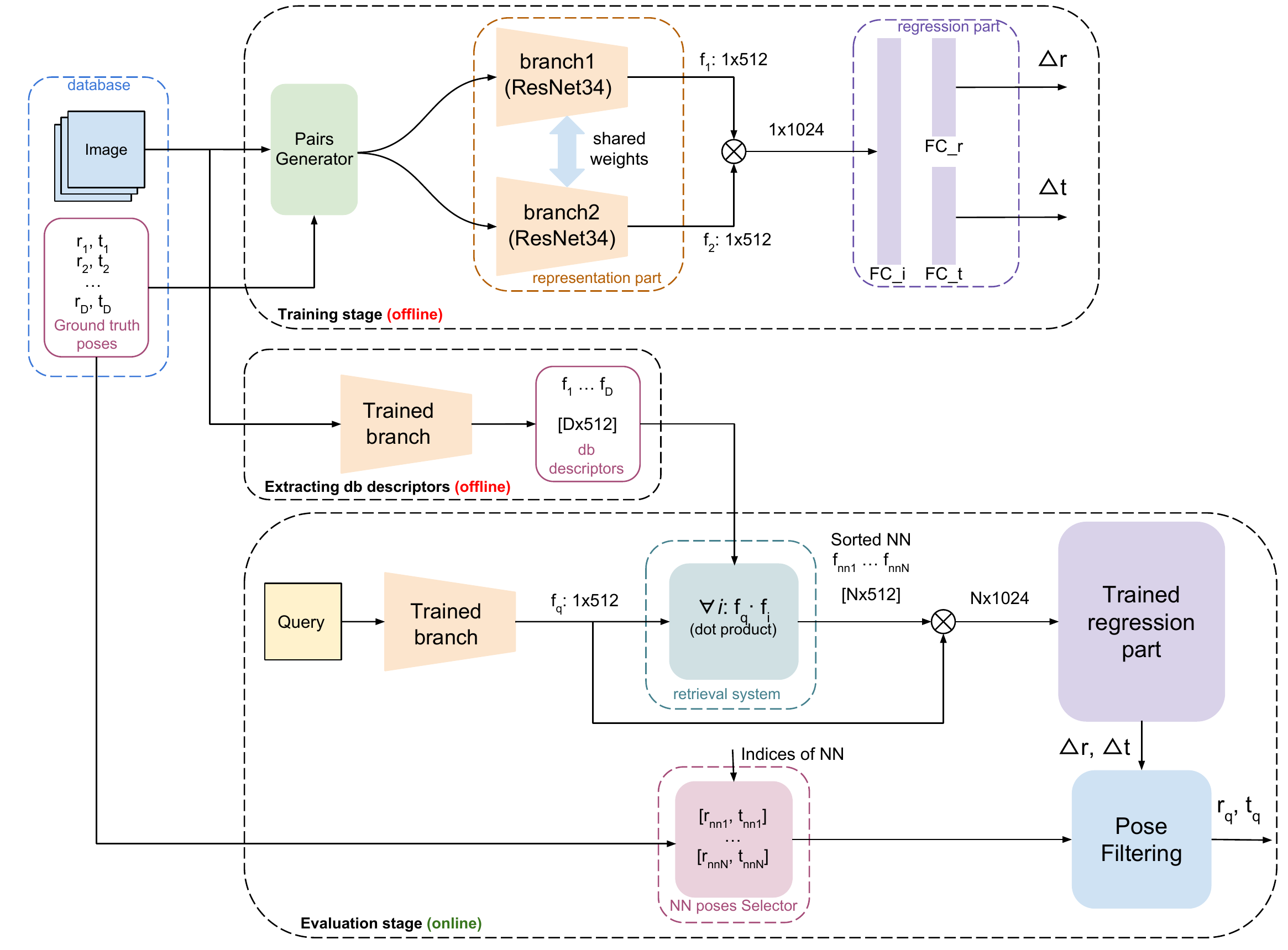}
		\caption{Pipeline of the proposed system. A Siamese based CNN network consisting of two branches pre-trained on ImageNet~\cite{ImageNet} is trained to directly regress relative pose between a pair of cameras (top). Once the training process is finished, we employ the branch as a descriptor to compute feature representations of database and query images. Then, the dot product is applied to these representations as a part of retrieval system and database descriptors are ranked according to higher similarity score. Consequently, query descriptor and its top N ranked database representations are concatenated and fed to the regression part of the network to predict pairwise relative pose. Finally, the proposed fusion algorithm naturally coalesces relative pose estimates and ground truth absolute poses to produce the full 6-DoF query location.}\label{fig:pipeline}
\end{figure*}

\section{Proposed approach}\label{sec:pipeline}
This section introduces the proposed localization approach for predicting camera pose. The method consists of two modules: a Siamese CNN network for relative pose computation and the localization pipeline. The input to the system is an RGB query image to be localized, and a database of images with their respective poses. At the first stage, we construct a set of training image pairs and use it to train a Siamese CNN to predict relative camera pose of each pair (Section~\ref{subsec:est_rel_pose}). It should be noted that the training image pairs can be independent of the scenes present in the localization database.Then, each trained branch of the network is considered a feature extractor and the extracted feature vectors can be utilized to identify the database images that are nearest neighbours (NN) to the query image in the feature space. Finally, relative pose estimates between the query and its neighbours are computed and then coalesced with ground truth absolute location of the corresponding database images in a novel fusion algorithm (Section~\ref{subsec:localization_system}) producing the full 6-DoF camera pose.

\subsection{Pairwise Pose Estimation}\label{subsec:est_rel_pose}

The first part of the proposed approach aims to directly estimate relative camera pose from an RGB image pair. The problem of regressing rigid body transformation between a pair of images has been well studied in recent years~\cite{deepHomography,MelekhovRelative,Demon}. Following~\cite{MelekhovRelative}, we construct a Siamese neural network with two representation branches connected to a common regression part as shown in Fig.~\ref{fig:pipeline}. The branches share the same set of weights and have ResNet34 architecture~\cite{ResNet} truncated at the last average pooling layer. The weights are initialised from a network pre-trained for large-scale image classification task~\cite{ImageNet}, and later fine-tuned for the relative pose estimation task as described below. The outputs of the representation branches are concatenated and passed through the regressor which consists of three fully-connected layers (FC), namely $FC_i$, $FC_r$ and $FC_t$, where the latter two predict relative orientation and translation, respectively. Fig.~\ref{fig:pipeline} shows a detailed description. The fully-connected layers are initialized randomly.

The output of the regression part is parameterized as 4-dimensional quaternion for relative orientation $ \Delta r $ and 3-dimensional $ \Delta t $ for relative translation ~\cite{MelekhovRelative,OliveiraRelative}. As the quaternions lie on a unit sphere, enforcing unit norm constraint on any 4-D vector outputs a valid rotation. Also the distance between two quaternions $s(r_i,r_k)$ can be measured by the Euclidean $l_2$ norm $||r_i - r_k||_2$, unlike other rotation parameterizations such as rotation matrices that lie on a manifold and distance computation requires finding an Euclidean embedding. For a more elaborate discussion we guide the reader to \cite{Hartley_IJCV13,Posenet2}. To regress relative camera pose, we train our CNN with the following objective criterion:


\begin{equation}\label{eq:eq_loss_function}
\mathcal{L} = \left\|\Delta t_{gt} - \Delta t\right\|_2 + \beta\left\|\Delta r_{gt} - \Delta r\right\|_2
\end{equation}
where $\Delta r_{gt}$ and $\Delta t_{gt}$ are the ground-truth relative orientation and translation respectively. To balance the loss for orientation and translation we use the parameter $\beta>0$ \cite{MelekhovRelative,OliveiraRelative}. The details of the training are described in Section \ref{sec:experiments}.


At test time, a pair of images is fed to the regression neural network, consisting of two branches, which directly estimates the real-valued parameters of the relative camera pose vector. Finally, the estimated quaternion and translation vectors are normalized to unit length. The normalized translation vector gives the translation direction from the database image to the query camera location. Although the translation vector predicted by our network contains scale information, we found that in practice recovering the scale using the approach discussed in Section~\ref{subsec:localization_system} is more accurate and reliable. 

\subsection{Localization Pipeline}\label{subsec:localization_system}
\noindent\textbf{Retrieving the nearest neighbours} In order to find the nearest database images to the query, 
it is important to obtain a suitable representation for both the query $q$ and the database images $D$.
Considering recent advances of CNN-based approaches in the field of image retrieval~\cite{retrievalGordo,retrievalRadenovich}, we use the fine-tuned network (one branch of the model architecture) from the first stage (Section~\ref{subsec:est_rel_pose}) as a feature extractor to encode the query and database images to fixed global representations (\ie 512 dimensional feature vectors, see Fig.~\ref{fig:pipeline}). In turn, the dot product of the query and database feature vectors is computed to obtain image similarity. Consequently, the database images are ranked according to similarity scores. Finally, the top $N$ ranked database images, $d = \{d^j| d^j \in D,j=1...N\}$ are selected as the nearest neighbours to the query image, $q$.

Although the retrieval stage is an important component in our pipeline, we adopted the simple approach above in this work and postpone deeper discussion and analysis to future work. The primary focus of our paper is not image retrieval, but camera localization. However, in Section~\ref{sec:experiments}, besides evaluating the performance of the complete pipeline, we also experimentally study how the system would perform with perfectly accurate retrieval stage. 

\begin{figure}[t!]
		\centering
        \includegraphics[width=0.3\textwidth]{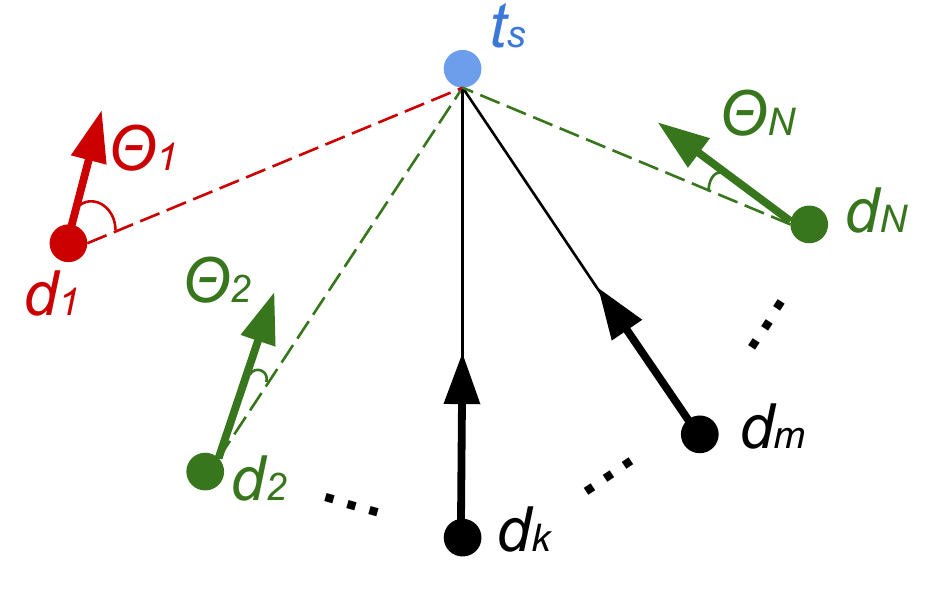}
		\caption{Estimation of query camera translation. 
        }\label{fig:hyp_filtering}
\end{figure}


\noindent\textbf{Compute Relative Pose}
In the next stage of the pipeline the Siamese CNN is used to regress relative camera pose for the image pairs $Q = q \times d$, $ \Delta R = \{ \Delta R^1, \Delta R^2...\Delta R^N \}$ and $ \Delta \hat{t} = \{\Delta \hat{t}^1, \Delta \hat{t}^2,... \Delta \hat{t}^N\} $. Here, $ \Delta R^j $ represents the relative orientation between the $j^{th}$ NN database image, $ d^j \in d $ and the query $q$. Similarly, $ \Delta \hat{t}^j $ is the relative translation direction.

\noindent\textbf{Pose hypotheses filtering}
The final part of the localization system is a novel fusion algorithm 
recovering the absolute query camera 
pose from these $N$ relative pose estimations. This process is illustrated schematically in Fig.~\ref{fig:hyp_filtering}.

From the shortlisted top ranked database images 
$d$, select a pair $p^s = \{ d^k, d^m \}$, where $p^s \subset d$ and $s = 1,2,\ldots, {{N}\choose{2}}$. Now, the translation direction predictions to the query $q$ from the images $ p^s $ are triangulated to obtain the location/translation parameter of query camera, $t^s$. This gives us ${{N}\choose{2}}$ hypotheses for the query translation, $\tilde{t}_q = \{t^s|s=1,2,\ldots, {{N}\choose{2}}\}$. Now, for each $t^s \in \tilde{t}_q$ the direction rays from the camera centers of the remaining images in $d$, $p^r = d$ \textbackslash $p^s$ to this triangulated camera location $t^s$ is obtained. If the direction ray associated with an image in $ p^r $ is within a pre-defined angular distance to the direction vector predicted by our network, then it is considered an inlier ($d_2$ and $d_N$ in Fig.~\ref{fig:hyp_filtering}) to the estimate $t^s$. The translation estimate $t^s \in \tilde{t}_q$, $s = 1,2,\ldots, {{N}\choose{2}}$ with the highest inlier count is assigned to the query camera. If two or more translation estimates have the same inlier count, then they are averaged to obtain the final query translation estimate.

Estimating rotation for the query camera is much simpler as the following equation can be used to obtain a hypothesis:
\begin{equation}
\Delta R^j = R_{j}^TR_{q}^j
\end{equation}
where $R_{j}$ is the orientation of the $j^{th}$ camera in $d$, $d^j \in d $ available as input to our system, and $R_{q}^j$ is the $j^{th}$ hypothesis of the query camera orientation. These results in $N$ hypotheses for query orientation, $\tilde{R}_q = \{R_{q}^1...R_{q}^N \}$. Instead of naively averaging the estimations, we apply a consensus based filtering similar to the process of estimating query translation. For each hypothesis, $R_{q}^j \in \tilde{R_q} $ a consensus set is formed from the remaining hypotheses in $\tilde{R_q}$ that lie within a pre-defined angular distance to $R_{q}^j $.
The number of elements in the consensus set is defined as the inlier count for $R_{q}^j $. The hypothesis with the highest inlier count is assigned as the orientation estimate for the query camera. When two or more hypotheses share the same inlier count, a robust rotation averaging algorithm \cite{Hartley_CVPR11} is applied to obtain the final query camera rotation.

\begin{figure*}[t!]
		\centering
		\includegraphics[width=0.9\textwidth]{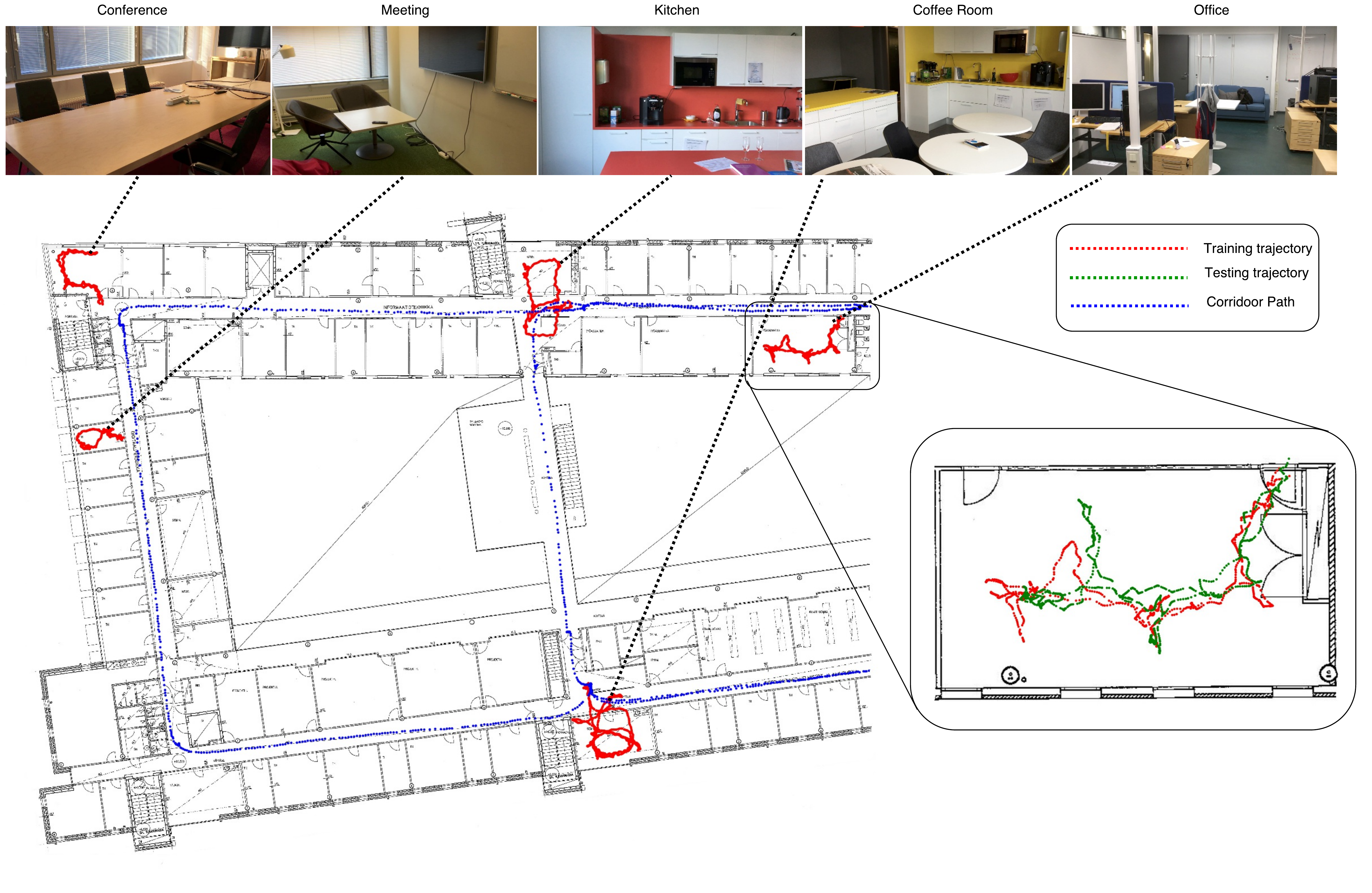}
		\caption{The University dataset. The proposed large-scale indoor localization dataset consists of 5 different scenes (segments) registered to a common global co-ordinate system. }\label{fig:university_dataset}
\end{figure*}

\section{Datasets}\label{sec:dataset}
We evaluate the proposed approach on two different dataset for indoor localization.

\noindent\textbf{7Scenes} Microsoft 7-Scenes dataset contains RGB-D images covering 7 different indoor locations~\cite{7ScenesDataset}. The dataset has been widely used for image-based localization~\cite{VidLoc,Posenet2,Posenet,HourglassPose} exhibiting significant variation in camera pose, motion blur and perceptual aliasing. In our experiments we utilize the same train and test datasets for each scene as provided in the original paper.

\noindent\textbf{University} The scenes in  \textit{7Scenes} dataset have their own coordinate system without being linked to each other in a global coordinate system.  Therefore, all existing machine learning models are trained and evaluated scene-wise. 
This fundamental limitation restricts to widely apply these approaches to real life scenarios where a large-scale environment consists of multiple sub-scenes.

We release a challenging indoor localization dataset, \textit{University}, with different locations that are all registered to a common global coordinate frame. For this paper, we consider a segment of the whole dataset, consisting of image sequences of 5 scenes, \textit{Office, Meeting, Kitchen, Conference, and Coffee Room}. The scenes are structured in a similar way to \textit{7Scenes}, with multiple traversals through each of the scenes.
The training and test split of the sequences are provided. Overall, the proposed dataset contains 9694 training and 5068 test images respectively. Some challenging test cases of \textit{University} dataset are presented in Figure 1 of the Supplementary material.

Ground-truth localization data of the dataset was generated using \textit{Google Project Tango's} tablet and high-resolution image sequences were collected by iPhone 6S mounted on top of the tablet. The device outputs two types of odometry estimations: Start of Service (SOS), which is the raw odometry and thus suffers from drift, and Area Learning (AL), which uses device's drift correction engine. As the AL engine fails sometimes~\cite{Laskar_ICIP16}, we use the odometry estimates from SOS and manually generated location constraints in a pose-graph optimization framework to generate a globally consistent map. 

\section{Experiments}\label{sec:experiments}
In this section we quantitatively demonstrate the effectiveness of the proposed system on \textit{7Scenes} and \textit{University} datasets. We compare our approach with the current state-of-the-art CNN-based methods, such as PoseNet~\cite{Posenet}, HourglassPose~\cite{HourglassPose}, LSTMPose~\cite{LSTMPose}, VidLoc~\cite{VidLoc}, and PoseNet with reprojection loss~\cite{Posenet2}, dubbed PoseNet2.


\begin{table*}[t!]
\begin{center}
\resizebox{1\textwidth}{!}{%
	\begin{tabular}{l | l | l l l l l | l l}
	\multirow{2}{*}{Scene} & \multicolumn{1}{c|}{Spatial} & \multicolumn{1}{c}{PoseNet}  & \multicolumn{1}{c}{LSTM-Pose} & \multicolumn{1}{c}{VidLoc} & \multicolumn{1}{c}{Hourglass-Pose} & \multicolumn{1}{c|}{PoseNet2} & \multicolumn{1}{c}{ResNet34-Pose} & \multicolumn{1}{c}{Ours} \\
    & \multicolumn{1}{c|}{Extent} & \multicolumn{1}{c}{\cite{Posenet}} & \multicolumn{1}{c}{\cite{LSTMPose}} & \multicolumn{1}{c}{\cite{VidLoc}} & \multicolumn{1}{c}{\cite{HourglassPose}} & \multicolumn{1}{c|}{\cite{Posenet2}} & \multicolumn{1}{c}{(baseline)} & \\
    \hline
    \hline
    Chess & $3\times 2 \times 1$m & $0.32$m, $8.12^{\circ}$ & $0.24$m, $5.77^{\circ}$ &  $0.18$m, N/A & $0.15$m, $6.53^{\circ}$ & $0.13$m, $4.48^{\circ}$ & $0.12$m, $6.69^{\circ}$ & $0.13$m, $6.46^{\circ}$\\
    Fire & $2.5\times 1 \times 1$m & $0.47$m, $14.4^{\circ}$ & $0.34$m, $11.9^{\circ}$ & $0.26$m, N/A & $0.27$m, $10.84^{\circ}$ & $0.27$m, $11.3^{\circ}$ & $0.31$m, $13.36^{\circ}$ & $0.26$m, $12.72^{\circ}$ \\
    Heads & $2\times 0.5 \times 1$m & $0.29$m, $12.0^{\circ}$ & $0.21$m, $13.7^{\circ}$ & $0.14$m, N/A & $0.19$m, $11.63^{\circ}$ & $0.17$m, $13.0^{\circ}$ & $0.16$m, $13.78^{\circ}$ & $0.14$m, $12.34^{\circ}$\\
    Office & $2.5\times 2 \times 1.5$m & $0.48$m, $7.68^{\circ}$ & $0.30$m, $8.08^{\circ}$ & $0.26$m, N/A & $0.21$m, $8.48^{\circ}$ & $0.19$m, $5.55^{\circ}$ & $0.21$m, $8.78^{\circ}$ & $0.21$m, $7.35^{\circ}$  \\
    Pumpkin & $2.5\times 2 \times 1$m & $0.47$m, $8.42^{\circ}$ & $0.33$m, $7.00^{\circ}$ & $0.36$m, N/A & $0.25$m, $7.01^{\circ}$ & $0.26$m, $4.75^{\circ}$ & $0.25$m, $7.89^{\circ}$ & $0.24$m, $6.35^{\circ}$ \\
    Red Kitchen & $4\times 3 \times 1.5$m & $0.59$m, $8.64^{\circ}$ & $0.37$m, $8.83^{\circ}$ & $0.31$m, N/A & $0.27$m, $10.15^{\circ}$ & $0.23$m, $5.35^{\circ}$ & $0.22$m, $9.35^{\circ}$ & $0.24$m, $8.03^{\circ}$\\
    Stairs & $2.5\times 2 \times 1.5$m & $0.47$m, $13.8^{\circ}$ & $0.40$m, $13.7^{\circ}$ & $0.26$m, N/A & $0.29$m, $12.46^{\circ}$ & $0.35$m, $12.4^{\circ}$ & $0.37$m, $14.45^{\circ}$ & $0.27$m, $11.82^{\circ}$ \\
    \hline \hline
    \multicolumn{2}{l|}{Average} & $0.44$m, $10.4^{\circ}$ & $0.31$m, $9.85^{\circ}$ & $0.25$m, N/A & $0.23$m, $9.53^{\circ}$ & $0.23$m, $8.12^{\circ}$ & $0.23$m, $10.61^{\circ}$ & $0.21$m, $9.30^{\circ}$
	\end{tabular}
    }
\end{center}
\vspace{-2mm}
\caption{Camera localization performance of the proposed method and existing RGB-only CNN-based approaches for \textit{7Scenes} datasets. We follow original notation presented in~\cite{Posenet} and provide median translation and orientation errors. In terms of localization error, our approach is superior to other methods utilizing similar loss (\ref{eq:eq_loss_function}) such as PoseNet~\cite{Posenet}, LSTM-Pose~\cite{LSTMPose}, VidLoc~\cite{VidLoc} and Hourglass-Pose~\cite{HourglassPose} for the all scenes. Furthermore, the proposed method outperforms PoseNet2~\cite{Posenet2} in 4 scenes and achieves better localization accuracy in general. However, it is important to note that both methods are not directly comparable, due to more advanced loss function optimized in~\cite{Posenet2}.}\label{lbl:table_results_7scenes} 
\end{table*}

\begin{table}[t!]
\begin{center}
\resizebox{.4\textwidth}{!}{%
  \begin{tabular}{l c | c c}
  
   Scene & Baseline & Proposed  \\ \hline \hline
   Office  & $2.76$m, $17.08^{\circ}$ & $0.57$m, $17.09^{\circ}$\\
   Meeting  & $2.13$m, $14.13^{\circ}$ & $2.30$m, $12.27^{\circ}$\\
   Kitchen  & $1.65$m, $12.92^{\circ}$ & $0.70$m, $11.72^{\circ}$\\
   Conference  & $4.97$ m, $17.18^{\circ}$ & $2.74$m, $15.00^{\circ}$\\
   \hline \hline
   Average &  $2.88$ m, $15.33^{\circ}$&  $1.58$ m, $14.02^{\circ}$ \\
 \end{tabular}
  }
\end{center}
\caption{Camera relocalization performance of the proposed approach and the baseline on \textit{University} dataset presented as median orientation and translation errors. Training is done using the training images of all scenes for both approaches. Evaluation is performed scene-wise.}\label{lbl:performance_on_university}
\end{table}

\begin{table}[t!]
\tiny
\begin{center}
\resizebox{.45\textwidth}{!}{%
  \begin{tabular}{l| c| c}
  
  \multicolumn{1}{l|}{Removed} & \multicolumn{2}{c}{Median error} \\
  \multicolumn{1}{l|}{scene} & \multicolumn{1}{c|}{position [m]} & \multicolumn{1}{c}{orientation [deg]}\\
  \hline\hline
   Chess & 0.27 & 13.05  \\ 
   Heads & 0.23 & 15.03  \\ 
   Red Kitchen & 0.36 & 12.60  \\ 
   
 \end{tabular}
  }
\end{center}
\vspace{-2mm}
\caption{Generalization performance of the proposed approach. Localization accuracy of the proposed method on unseen scenes of \textit{7Scenes} dataset.}\label{tbl:generalization_removed_scenes}
\end{table}

\begin{table}[t!]
\tiny
\begin{center}
\resizebox{.45\textwidth}{!}{%
  \begin{tabular}{l| c| c}
  
  \multirow{2}{*}{Scene} & \multicolumn{2}{c}{Median error} \\
                          & \multicolumn{1}{c|}{position [m]} & \multicolumn{1}{c}{orientation [deg]}\\
  \hline\hline
   Chess       & 0.31 & 15.05  \\ 
   Fire        & 0.40 & 19.00  \\ 
   Heads       & 0.24 & 22.15  \\ 
   Office      & 0.38 & 14.14  \\ 
   Pumpkin     & 0.44 & 18.24  \\ 
   Red Kitchen & 0.41 & 16.51  \\ 
   Stairs      & 0.35 & 23.55  \\ 
   \hline \hline
   Average     & 0.36 & 18.38
 \end{tabular}
  }
\end{center}
\vspace{-2mm}
\caption{Generalization performance of the proposed approach. The network is trained on only \textit{University} and evaluated on \textit{7Scenes} dataset.}\label{tbl:generalization_trained_on_aalto}
\end{table}

According to Fig.\ref{fig:pipeline}, representation part of the proposed system is based on a Siamese architecture. In this work, we initialize our model using original ResNet34 model truncated at the last average pooling layer and pre-trained on ImageNet~\cite{ImageNet} as a structure of each branch.

\noindent\textbf{Training data} We start experiments by evaluating the system on \textit{7Scenes} dataset. As it is mentioned in Section~\ref{sec:dataset}, the dataset consists of different indoor scenes and each of them provides training and testing image sequences. Since the system requires an image pair to learn relative pose, the following strategy is applied to obtain training dataset. For every image in the training set of each scene, we find a corresponding image from the same set so that they have sufficiently overlapping field of view. Total number of training pairs of a scene is equal to the number of images in the training sequence of this scene in the original dataset.  
Finally, the training pairs from all the scenes are merged into a single training set for training the CNN in our approach.

\noindent\textbf{Training details} As a preprocessing step, the training images are resized to 256 pixels on the smaller side while maintaining the aspect ratio of the resized image. The training images are further mean-centered and normalized using standard deviation computed over the whole training set. To ensure fixed sized input to our network, we use random crops (224 $ \times$ 224) during training and perform the evaluation using central crops at test time. The network is trained for 200 epochs with an initial learning rate of 0.1 which is gradually decreased by 10 times after every 50 epochs. The loss (\ref{eq:eq_loss_function}) is minimized using stochastic gradient descent with a batch size of 64 training samples. The scale factor $\beta$ is set to 1 after empirical evaluation for all our experiments. The weight decay is set to $10^{-5}$ and no dropout was used in our experiments. All experiments were evaluated on two NVIDIA Titan X GPUs using Torch7~\cite{Torch7} machine learning framework.

\noindent\textbf{Evaluation stage}
The input to the system is a database containing the list of images from the training set of all scenes (for a given dataset) and their respective camera poses. The combined list of test images from all scenes constitute the query set. For each query, we retrieve its top 5 NN ($N$ = 5) from the database images using neural representations from our trained representation branch.


The query and its NN are then fed sequentially to our Siamese model to obtain the relative camera pose estimates. From a practical standpoint it is not necessary to feed the query and its NN images through the full network model. The only component of our network that requires pairwise input is the regression part, which takes in input from the representation part of each branch of the Siamese model. Also, both the representation branches share the same parameters and the output of the representation part is already used in the first stage of our pipeline to compute image similarity. Thereby, to compute relative pose, we simply feed the representations of the query and its NN in a pairwise manner to the regression component. 

The relative pose estimations are then robustly fused to obtain the query camera pose. The angular distance threshold of inliers for both translation and rotation is 20 degrees. 

\subsection{Quantitative Results}\label{ssec:results}
We compare our proposed system with the existing CNN based localization methods on~\textit{7Scenes}, while for~\textit{University} dataset we provide an evaluation of our proposed system and a baseline method. The results are shown in Table~\ref{lbl:table_results_7scenes} and Table~\ref{lbl:performance_on_university}. For both datasets, the localization performance is measured as the median orientation and translation error over each scene. 

For several scenes in the \textit{7Scenes} dataset we outperform other CNN-based methods in camera relocalization. 
In particular, we perform favourably  to the current best performing method, PoseNet2 on several scenes. However, PoseNet2 is not directly comparable to our work as it uses a more sophisticated loss function during training and a different CNN architecture. 

For a fairer evaluation we compare our system with a baseline model consisting of pre-trained convolutional layers of ResNet34 architecture and a regression part replicating the one utilized in the proposed approach (but without the Siamese architecture). We entitle this model {ResNet34-Pose}. Following~\cite{Posenet2,HourglassPose}, the baseline is trained and evaluated scene-wise. Table~\ref{lbl:table_results_7scenes} shows that our proposed system has a consistent improvement over the baseline for both rotation and translation across all the scenes. Although the margin of improvement is not large, it is to be noted that all the existing methods (including the baseline) are trained in a scene-specific manner whereas our system was designed to inherently overcome this fundamental limitation and allows us to train and test our model jointly on all the scenes. That is, our approach uses the same network for all 7 scenes whereas other approaches of Table \ref{lbl:table_results_7scenes} have one network per scene (\eg in total ResNet34-Pose has thus 7 times more parameters that are learnt).

The performance increase of our system compared to the baseline can be attributed to a number of factors: \emph{i)} representation sharing across scenes during training, \emph{ii)} generating multiple hypothesis for query camera pose followed by robust pose filtering, \emph{iii)} larger training set.  These factors, although plausible have not been experimentally validated in this paper and we leave it for future work. 

For the \textit{University} dataset, our system and the baseline are trained jointly on the scenes: \textit{Office, Meeting, Kitchen} and \textit{Conference}. The scene \textit{Coffee Room} is a recent addition to the database, and due to time constraints we could not train and evaluate our proposed system and {ResNet34-Pose} on this scene. However, we use it to evaluate our trained model in Section \ref{sec:generalize}. The performance evaluation is presented in Table~\ref{lbl:performance_on_university}. The results show that the margin of translational error between the baseline and our system has increased significantly. In particular, it has increased from 2cm in \textit{7 Scenes} (all scenes combined) to 130cm in \textit{University}. Although it does demonstrate that the proposed system performs better even under similar training setup, it also provides additional insight on the scalability of our system. As mentioned in Section~\ref{sec:dataset}, the \textit{University} dataset consists of several scenes spread over an area of 2500 $m^2$. Absolute pose prediction models like the baseline model need to maintain a track of the spread or scale of the map, while models like our system are not much affected by scale. Our system essentially removes the influence of scale by finding the NN and solving the relative pose problem which does not depend on the scale of the map/dataset.

\vspace{-1mm}
\subsection{Generalization Performance} \label{sec:generalize}
\vspace{-1mm}

Current machine learning models for camera localization are not only restricted to scene-wise training and evaluation, but also limited in their applicability to previously unseen scenes. In this section we experimentally demonstrate the generalization capability of our pipeline to data previously unseen during training. 

We hold out one of the scenes in \textit{7Scenes} dataset for evaluation and train our model on the remaining 6 scenes. In particular, we held out \textit{Chess, Heads,} and \textit{RedKitchen} separately as evaluation sets. Table~\ref{tbl:generalization_removed_scenes} shows a graceful drop in performance on the held out test scene compared to the case where our model was trained on all the 7 scenes (Table \ref{lbl:table_results_7scenes}). In \textit{University} dataset, we evaluate on \textit{Coffee Room} using the model trained on the remaining 4 scenes. The median position and orientation error were 1.44 m and 19.22 degrees. 

We further evaluate the performance on the \textit{7Scenes} dataset using our model trained on the \textit{University}  dataset (excluding \textit{Coffee Room}). According to Table~\ref{tbl:generalization_trained_on_aalto}, the performance drop is not drastic, with the mean of the median error over all the scenes drop by 9 degrees and 15 cm for rotation and translation respectively.

\begin{table}[t!]
 \begin{center}
 \resizebox{.5\textwidth}{!}{%
 	\begin{tabular}{l | l l l }
 	Scene  & Viewpoint 0  & Viewpoint 3 & Viewpoint 7 \\
     \hline
     \hline
     Chess  & $0.19$m, $7.48^{\circ}$ & $0.16$m, $7.26^{\circ}$ &  $0.16$m, $7.61^{\circ}$ \\
     Fire   & $0.13$m, $6.61^{\circ}$ & $0.10$m, $6.45^{\circ}$ & $0.11$m, $6.32^{\circ}$  \\
     Heads  & $0.25$m, $8.74^{\circ}$ & $0.25$m, $8.54^{\circ}$ & $0.25$m, $8.71^{\circ}$ \\
     Office & $0.21$m, $11.13^{\circ}$ & $0.19$m, $11.14^{\circ}$ & $0.19$m, $11.95^{\circ}$ \\
     Pumpkin  & $0.24$m, $9.39^{\circ}$ & $0.26$m, $9.50^{\circ}$ & $0.25$m, $9.35^{\circ}$ \\
     Red Kitchen  & $0.21$m, $7.59^{\circ}$ & $0.19$m, $7.41^{\circ}$ & $0.19$m, $7.45^{\circ}$ \\
     Stairs  & $0.23$m, $7.92^{\circ}$ & $0.23$m, $8.26^{\circ}$ & $0.23$m, $8.56^{\circ}$ \\
     \hline \hline
     Average & $0.21$m, $8.41^{\circ}$ & $0.20$m, $8.37^{\circ}$ & $0.20$m, $8.44^{\circ}$
 	\end{tabular}
     }
 \end{center}
 \vspace{-4mm}
 \caption{Camera relocalization accuracy of the proposed system for different viewpoint changes between the query and the database image.}\label{lbl:results_perfect_retrieval}
 \vspace{-1mm}
 \end{table}

\vspace{-1mm}
\subsection{Ideal Retrieval Results}\label{subsec:perfect_retrieval}
We now evaluate the scenario when the retrieval stage of our pipeline returns only the true nearest neighbours to the query. We further evaluate the effect of image spacing between the query and the retrieved NN. These experiments are evaluated on \textit{7Scenes} dataset. 

Due to low image spacing between training images across all the scenes in \textit{7Scenes}, a query often has more than 300 true NN. Now, for a given query we use ground truth to sort all the training images from the corresponding scene using a metric similar to \eqref{eq:eq_loss_function}. We then create a sublist consisting of the top 355 images from this sorted list. From this sublist, we further select 8 sets of 5 images each at an interval of 50 ranks. That is, the first set (Viewpoint 0 in Table \ref{lbl:results_perfect_retrieval}) contains images ranked 1-5, the next set (Viewpoint 1 in Supplementary) consisting of images ranked 51-55 and so on.  
We then evaluate our proposed system using these sets of true NN instead of the one obtained using neural representations. Table \ref{lbl:results_perfect_retrieval} (and Table in Supplementary) shows that the proposed system has a consistent performance across wide viewpoint variation in the true NN. This is an indication that the pipeline is robust to the quality of the nearest neighbours. They do not necessarily need to be the database images that have most overlap with the query. On the other hand, the result also shows that with \textit{7Scenes} our choice of $N=5$ might not be the optimal choice. Increasing $N$ will increase the chances of retrieving the true NN, and the consistent performance across all scenes and viewpoint changes suggests that the true NN have a higher likelihood of forming a consensus set (\cf Section \ref{subsec:localization_system}).

\vspace{-1mm}
\section{Conclusion} \label{sec:conclusion}
We addressed some of the challenges and limitations of the current setup in which machine learning models are trained and evaluated for camera localization. By leveraging the training images both at training and test time, we are able to mitigate these limitations and achieve competitive results on challenging datasets. Results demonstrate that the scope of the proposed system is easily extendable to scenes without prior training. 

As future work, possible directions include training the network simultaneously with both relative pose and image similarity objectives \cite{zamir2016generic}. Also, learning a better generic relative camera pose estimator~\cite{Demon} can improve the generalization performance of the proposed system.

 {\small
 \bibliographystyle{ieee}
 \bibliography{egbib}

\begin{thebibliography}{10}\itemsep=-1pt

\bibitem{ImageRetrievalBabenko}
A.~Babenko, A.~Slesarev, A.~Chigorin, and V.~S. Lempitsky.
\newblock Neural codes for image retrieval.
\newblock In {\em Computer Vision - {ECCV}}, 2014.

\bibitem{SURF}
H.~Bay, A.~Ess, T.~Tuytelaars, and L.~Van~Gool.
\newblock Speeded-up robust features {(SURF)}.
\newblock {\em Comput. Vis. Image Underst.}, 110(3):346--359, 2008.

\bibitem{DSAC}
E.~Brachmann, A.~Krull, S.~Nowozin, J.~Shotton, F.~Michel, S.~Gumhold, and
  C.~Rother.
\newblock {DSAC} - differentiable {RANSAC} for camera localization.
\newblock {\em CoRR}, abs/1611.05705, 2016.

\bibitem{PnPBujnak}
M.~Bujnak, Z.~Kukelova, and T.~Pajdla.
\newblock New efficient solution to the absolute pose problem for camera with
  unknown focal length and radial distortion.
\newblock In {\em {ACCV}}, 2011.

\bibitem{VidLoc}
R.~Clark, S.~Wang, N.~T. Andrew~Markham, and H.~Wen.
\newblock Vid{L}oc: {A} deep spatio-temporal model for 6-{D}o{F} video-clip
  relocalization.
\newblock In {\em {CVPR}}, 2017.

\bibitem{Torch7}
R.~Collobert, K.~Kavukcuoglu, and C.~Farabet.
\newblock Torch7: A matlab-like environment for machine learning.
\newblock In {\em BigLearn, NIPS Workshop}, 2011.

\bibitem{BOWCula}
O.~G. Cula and K.~J. Dana.
\newblock Compact representation of bidirectional texture functions.
\newblock In {\em {CVPR}}, 2001.

\bibitem{ImageNet}
J.~Deng, W.~Dong, R.~Socher, L.-J. Li, K.~Li, and L.~Fei-Fei.
\newblock {Image{N}et: A Large-Scale Hierarchical Image Database}.
\newblock In {\em CVPR}, 2009.

\bibitem{deepHomography}
D.~DeTone, T.~Malisiewicz, and A.~Rabinovich.
\newblock Deep image homography estimation.
\newblock {\em CoRR}, abs/1606.03798, 2016.

\bibitem{retrievalGordo}
A.~Gordo, J.~Almaz{\'{a}}n, J.~Revaud, and D.~Larlus.
\newblock Deep image retrieval: Learning global representations for image
  search.
\newblock {\em CoRR}, abs/1604.01325, 2016.

\bibitem{Hartley_CVPR11}
R.~Hartley, K.~Aftab, and J.~Trumpf.
\newblock L1 rotation averaging using the weiszfeld algorithm.
\newblock In {\em Computer Vision and Pattern Recognition (CVPR), 2011 IEEE
  Conference on}, pages 3041--3048. IEEE, 2011.

\bibitem{Hartley_IJCV13}
R.~Hartley, J.~Trumpf, Y.~Dai, and H.~Li.
\newblock Rotation averaging.
\newblock {\em International journal of computer vision}, 103(3):267--305,
  2013.

\bibitem{ResNet}
K.~He, X.~Zhang, S.~Ren, and J.~Sun.
\newblock Deep residual learning for image recognition.
\newblock In {\em {CVPR}}, 2016.

\bibitem{SegmentationHong}
S.~Hong, H.~Noh, and B.~Han.
\newblock Decoupled deep neural network for semi-supervised semantic
  segmentation.
\newblock In {\em NIPS}, 2015.

\bibitem{Posenet2}
A.~Kendall and R.~Cipolla.
\newblock Geometric loss functions for camera pose regression with deep
  learning.
\newblock {\em CoRR}, abs/1704.00390, 2017.

\bibitem{Posenet}
A.~Kendall, M.~Grimes, and R.~Cipolla.
\newblock Pose{N}et: {A} convolutional network for real-time 6-dof camera
  relocalization.
\newblock In {\em {ICCV}}, 2015.

\bibitem{ImagenetKrizh}
A.~Krizhevsky, I.~Sutskever, and G.~E. Hinton.
\newblock Imagenet classification with deep convolutional neural networks.
\newblock In {\em Advances in NIPS}, 2012.

\bibitem{Laskar_ICIP16}
Z.~Laskar, S.~Huttunen, D.~Herrera, E.~Rahtu, and J.~Kannala.
\newblock Robust loop closures for scene reconstruction by combining odometry
  and visual correspondences.
\newblock In {\em Image Processing (ICIP), 2016 IEEE International Conference
  on}, pages 2603--2607. IEEE, 2016.

\bibitem{PrioritizedSearchSnavely}
Y.~Li, N.~Snavely, and D.~P. Huttenlocher.
\newblock Location recognition using prioritized feature matching.
\newblock In {\em Proc. {ECCV}}, 2010.

\bibitem{SIFT}
D.~G. Lowe.
\newblock Distinctive image features from scale-invariant keypoints.
\newblock {\em Int. J. Comput. Vision}, 60(2):91--110, Nov. 2004.

\bibitem{HourglassPose}
I.~Melekhov, J.~Ylioinas, J.~Kannala, and E.~Rahtu.
\newblock Image-based localization using hourglass networks.
\newblock {\em CoRR}, abs/1703.07971, 2017.

\bibitem{MelekhovRelative}
I.~Melekhov, J.~Ylioinas, J.~Kannala, and E.~Rahtu.
\newblock Relative camera pose estimation using convolutional neural networks.
\newblock {\em CoRR}, abs/1702.01381, 2017.

\bibitem{SegmentationNoh}
H.~Noh, S.~Hong, and B.~Han.
\newblock Learning deconvolution network for semantic segmentation.
\newblock In {\em {ICCV}}, 2015.

\bibitem{OliveiraRelative}
G.~L. Oliveira, N.~Radwan, W.~Burgard, and T.~Brox.
\newblock Topometric localization with deep learning.
\newblock {\em CoRR}, abs/1706.08775, 2017.

\bibitem{Oxford5k}
J.~Philbin, O.~Chum, M.~Isard, J.~Sivic, and A.~Zisserman.
\newblock Object retrieval with large vocabularies and fast spatial matching.
\newblock In {\em Proc.{CVPR}}, 2007.

\bibitem{retrievalRadenovich}
F.~Radenovi{\'c}, G.~Tolias, and O.~Chum.
\newblock {CNN} image retrieval learns from {BoW}: Unsupervised fine-tuning
  with hard examples.
\newblock In {\em ECCV}, 2016.

\bibitem{ORB}
E.~Rublee, V.~Rabaud, K.~Konolige, and G.~Bradski.
\newblock Orb: An efficient alternative to sift or surf.
\newblock In {\em Proc. {ICCV}}, 2011.

\bibitem{SattlerTopDbSearch2}
T.~Sattler, M.~Havlena, F.~Radenovic, K.~Schindler, and M.~Pollefeys.
\newblock Hyperpoints and fine vocabularies for large-scale location
  recognition.
\newblock In {\em {ICCV}}, 2015.

\bibitem{PrioritizedSearchSattler}
T.~Sattler, B.~Leibe, and L.~Kobbelt.
\newblock Efficient effective prioritized matching for large-scale image-based
  localization.
\newblock {\em IEEE TPAMI}, 2016.

\bibitem{sattler_CVPR2017}
T.~Sattler, A.~Torii, J.~Sivic, M.~Pollefeys, H.~Taira, M.~Okutomi, and
  T.~Pajdla.
\newblock {Are Large-Scale 3D Models Really Necessary for Accurate Visual
  Localization?}
\newblock In {\em {CVPR}}, 2017.

\bibitem{SattlerTopDbSearch1}
T.~Sattler, T.~Weyand, B.~Leibe, and L.~Kobbelt.
\newblock Image retrieval for image-based localization revisited.
\newblock In {\em {BMVC}}, 2012.

\bibitem{scorfShotton}
J.~Shotton, B.~Glocker, C.~Zach, S.~Izadi, A.~Criminisi, and A.~Fitzgibbon.
\newblock Scene coordinate regression forests for camera relocalization in
  {RGB-D} images.
\newblock In {\em {CVPR}}, 2013.

\bibitem{7ScenesDataset}
J.~Shotton, B.~Glocker, C.~Zach, S.~Izadi, A.~Criminisi, and A.~Fitzgibbon.
\newblock Scene coordinate regression forests for camera relocalization in
  {RGB-D} images.
\newblock In {\em {CVPR}}. IEEE, 2013.

\bibitem{GoogLeNet}
C.~Szegedy, W.~Liu, Y.~Jia, P.~Sermanet, S.~E. Reed, D.~Anguelov, D.~Erhan,
  V.~Vanhoucke, and A.~Rabinovich.
\newblock Going deeper with convolutions.
\newblock In {\em {CVPR}}, 2015.

\bibitem{Demon}
B.~Ummenhofer, H.~Zhou, J.~Uhrig, N.~Mayer, E.~Ilg, A.~Dosovitskiy, and
  T.~Brox.
\newblock De{M}o{N}: Depth and motion network for learning monocular stereo.
\newblock In {\em Proc. {CVPR}}. IEEE, 2017.

\bibitem{scorfValentin}
J.~Valentin, M.~Nie{\ss}ner, J.~Shotton, A.~Fitzgibbon, S.~Izadi, and P.~H.~S.
  Torr.
\newblock Exploiting uncertainty in regression forests for accurate camera
  relocalization.
\newblock In {\em {CVPR}}, 2015.

\bibitem{BOWVarma}
M.~Varma and A.~Zisserman.
\newblock Classifying images of materials: Achieving viewpoint and illumination
  independence.
\newblock In {\em {ECCV}}, 2002.

\bibitem{LSTMPose}
F.~Walch, C.~Hazirbas, L.~Leal{-}Taix{\'{e}}, T.~Sattler, S.~Hilsenbeck, and
  D.~Cremers.
\newblock Image-based localization with spatial {LSTM}s.
\newblock {\em CoRR}, abs/1611.07890, 2016.

\bibitem{zamir2016generic}
A.~R. Zamir, T.~Wekel, P.~Agrawal, C.~Wei, J.~Malik, and S.~Savarese.
\newblock Generic 3d representation via pose estimation and matching.
\newblock In {\em European Conference on Computer Vision}, pages 535--553.
  Springer International Publishing, 2016.

\end{thebibliography}
 }

\clearpage
\onecolumn
\null
\vskip .375in
   \begin{center}
      {\Large \bf Camera Relocalization by Computing Pairwise Relative Poses \\Using Convolutional Neural Network \\ - Supplementary Material - \par}
      \vspace*{24pt}
      \vskip .5em
      \vspace*{12pt}
   \end{center}

\begin{figure*}[h!]
		\centering
		\begin{subfigure}[t]{.5\textwidth}
            \centering
		    \includegraphics[width=\textwidth]{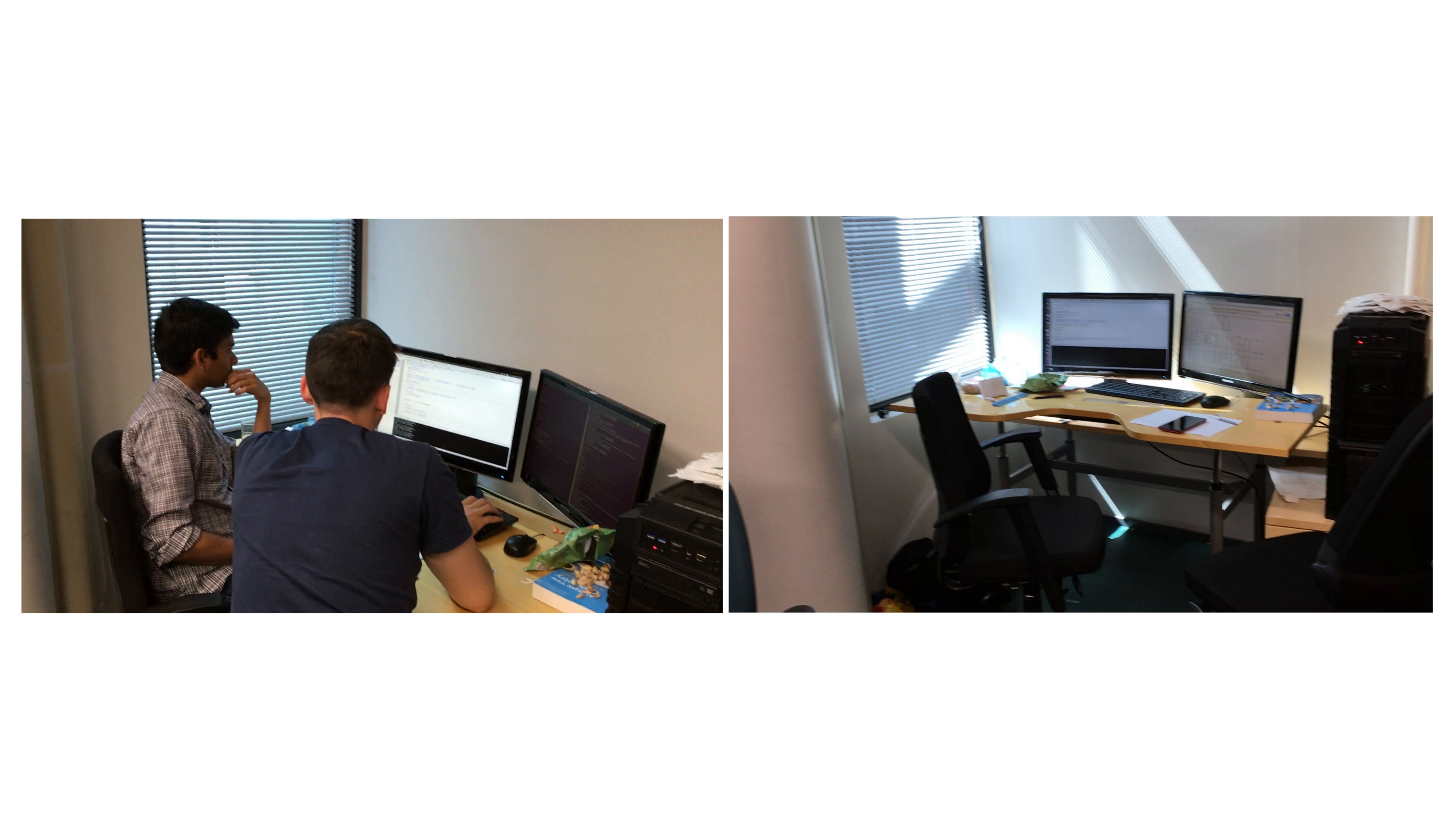}
            \caption{Changing environment}
        \end{subfigure}%
        ~ 
        \begin{subfigure}[t]{.5\textwidth}
            \centering
            \includegraphics[width=\textwidth]{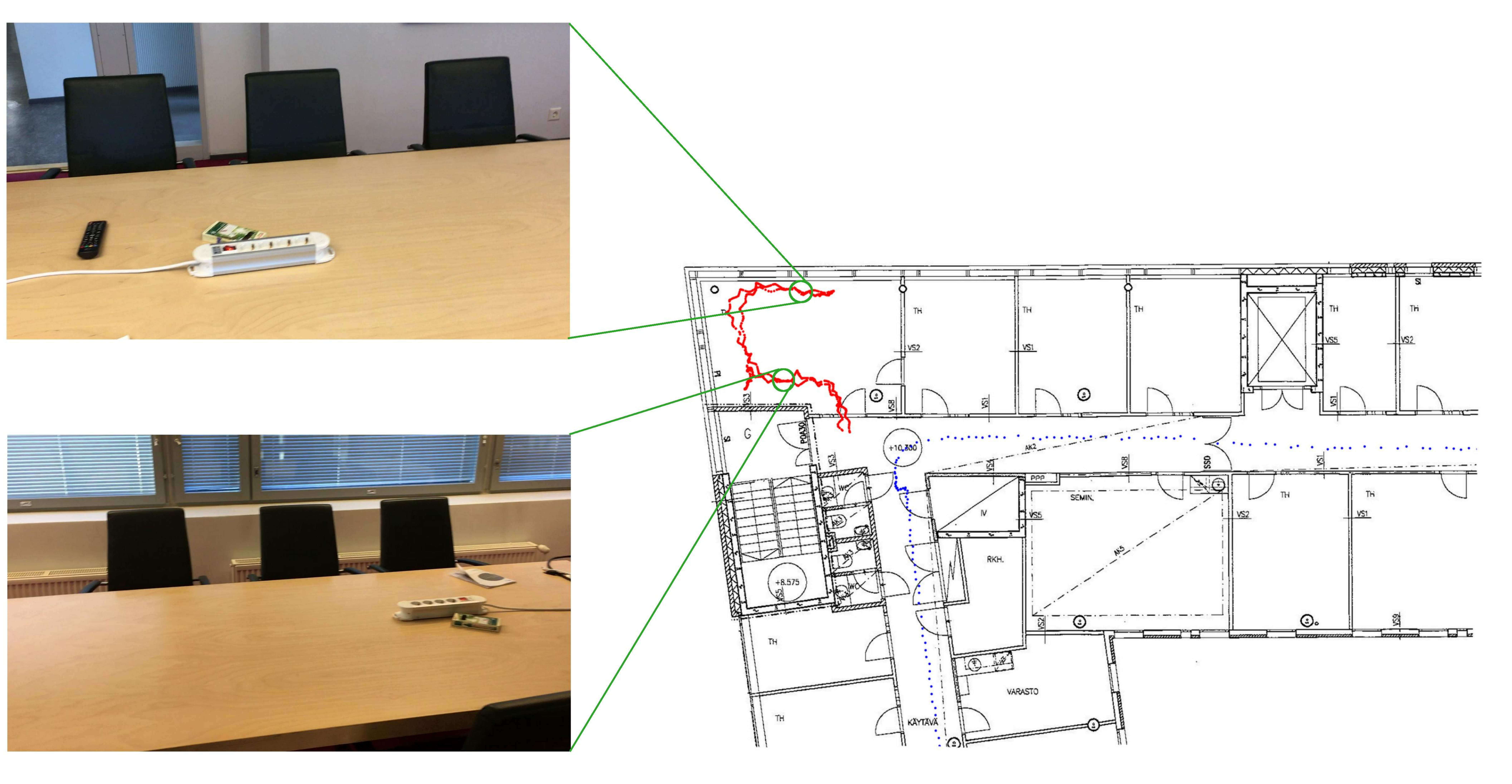}
            \caption{Perceptual aliasing}
        \end{subfigure}
\caption{Some challenging cases in \textit{University} dataset.}
\end{figure*}

\begin{table*}[h!]
 \begin{center}
 \resizebox{.9\textwidth}{!}{%
 	\begin{tabular}{l | l l l l l}
 	Scene  & Viewpoint 1  & Viewpoint 2 & Viewpoint 4 &  Viewpoint 5 & Viewpoint 6\\
     \hline
     \hline
     Chess  & $0.17$m, $7.50^{\circ}$ & $0.16$m, $7.30^{\circ}$ &  $0.17$m, $7.36^{\circ}$ &  $0.17$m, $7.30^{\circ}$ &  $0.16$m, $7.59^{\circ}$\\
     Fire   & $0.10$m, $6.37^{\circ}$ & $0.11$m, $6.29^{\circ}$ & $0.10$m, $6.44^{\circ}$ &  $0.10$m, $6.32^{\circ}$  &  $0.11$m, $6.44^{\circ}$\\
     Heads  & $0.24$m, $8.50^{\circ}$ & $0.25$m, $8.82^{\circ}$ & $0.24$m, $8.58^{\circ}$ &  $0.24$m, $8.46^{\circ}$ &  $0.25$m, $8.83^{\circ}$\\
     Office & $0.19$m, $11.13^{\circ}$ & $0.19$m, $11.16^{\circ}$ & $0.19$m, $11.11^{\circ}$ &  $0.20$m, $11.13^{\circ}$ &  $0.19$m, $11.06^{\circ}$\\
     Pumpkin  & $0.26$m, $8.95^{\circ}$ & $0.26$m, $9.27^{\circ}$ & $0.26$m, $9.45^{\circ}$ &  $0.26$m, $9.31^{\circ}$ &  $0.25$m, $9.31^{\circ}$ \\
     Red Kitchen  & $0.20$m, $7.40^{\circ}$ & $0.20$m, $7.45^{\circ}$ & $0.19$m, $7.48^{\circ}$ &  $0.19$m, $7.48^{\circ}$ &  $0.19$m, $7.47^{\circ}$ \\
     Stairs  & $0.22$m, $8.80^{\circ}$ & $0.23$m, $8.15^{\circ}$ & $0.23$m, $8.36^{\circ}$ &  $0.23$m, $8.30^{\circ}$ &  $0.23$m, $8.37^{\circ}$\\
     \hline \hline
     Average & $0.21$m, $8.38^{\circ}$ & $0.20$m, $8.35^{\circ}$ & $0.20$m, $8.40^{\circ}$ &  $0.20$m, $8.33^{\circ}$ &  $0.20$m, $8.44^{\circ}$ 
 	\end{tabular}
     }
 \end{center}
 \caption{Camera relocalization accuracy of the proposed system for different viewpoint changes between the query and the database image. The explanation of \textbf{Viewpoint N} notation is provided Section~\ref{subsec:perfect_retrieval}}
 \end{table*}
 
 \begin{figure*}[t!]
 	\centering
 	\begin{subfigure}[t]{.5\textwidth}
 		\centering
 		\includegraphics[width=\textwidth]{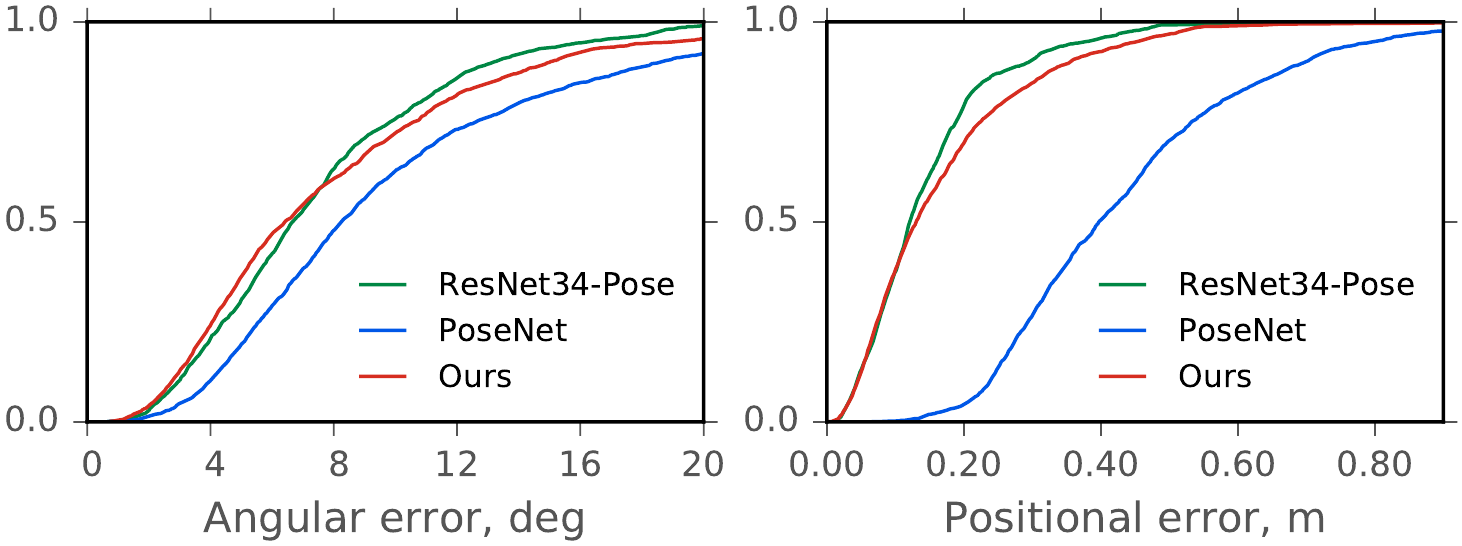}
 		\caption{Chess}
 	\end{subfigure}%
 	~
 	\begin{subfigure}[t]{.5\textwidth}
 		\centering
 		\includegraphics[width=\textwidth]{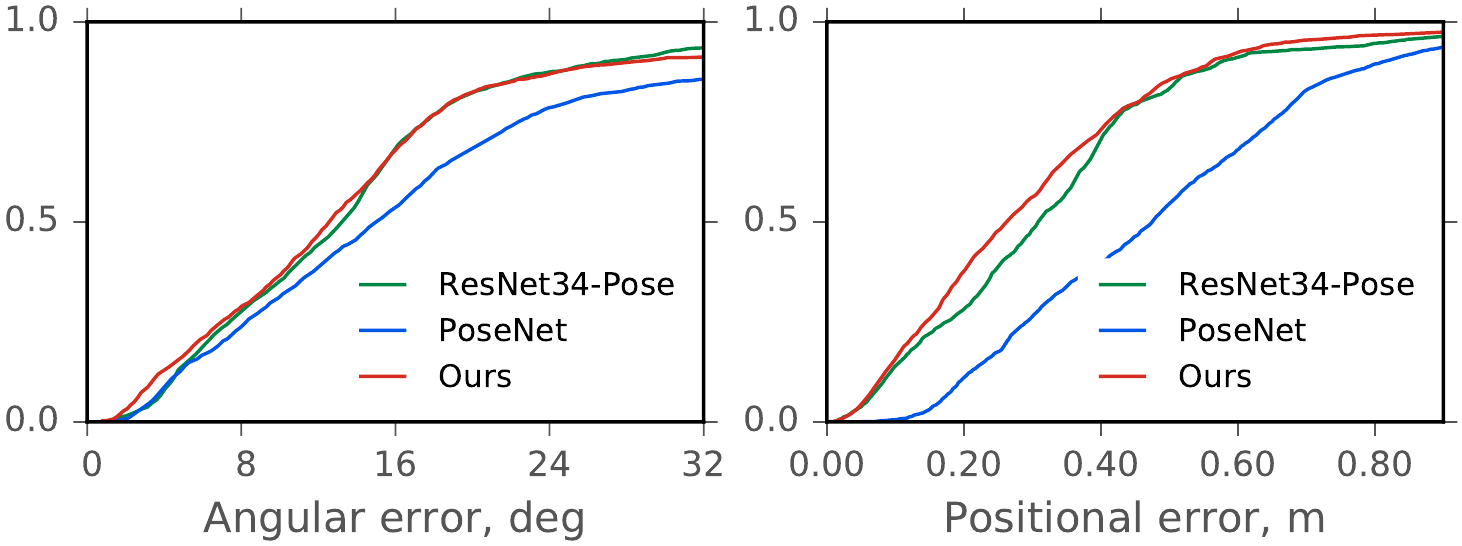}
 		\caption{Fire}
 	\end{subfigure}
 	~
 	\begin{subfigure}[t]{.5\textwidth}
 		\centering
 		\includegraphics[width=\textwidth]{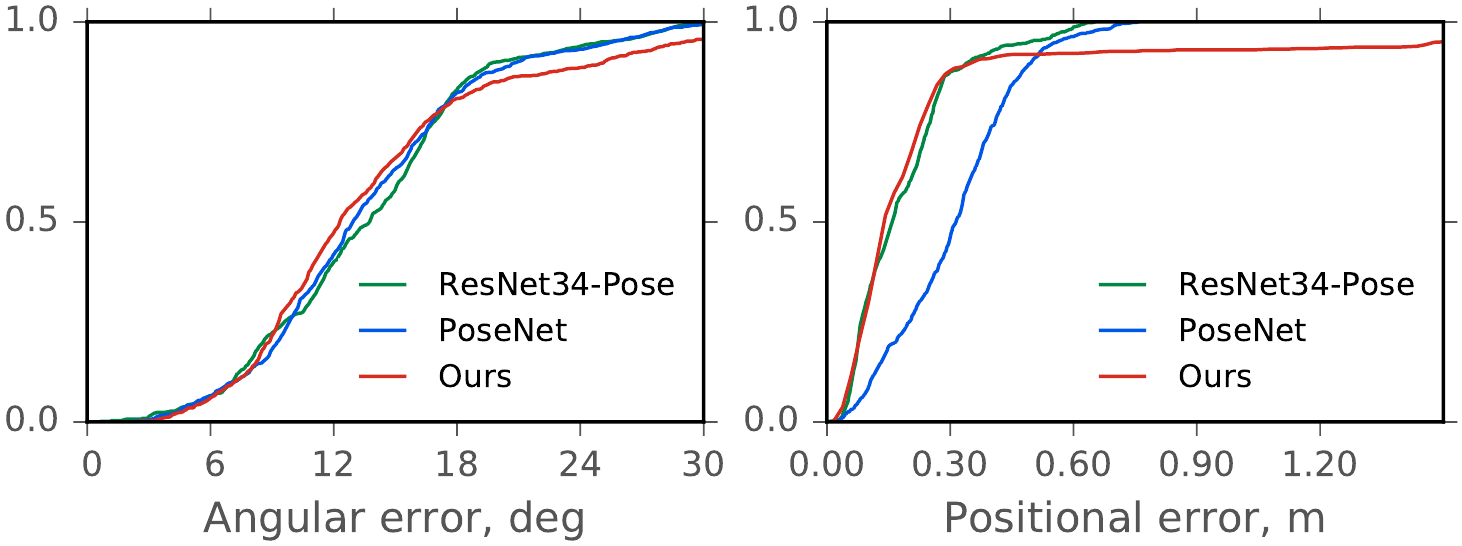}
 		\caption{Heads}
 	\end{subfigure}%
 	~
 	\begin{subfigure}[t]{.5\textwidth}
 		\centering
 		\includegraphics[width=\textwidth]{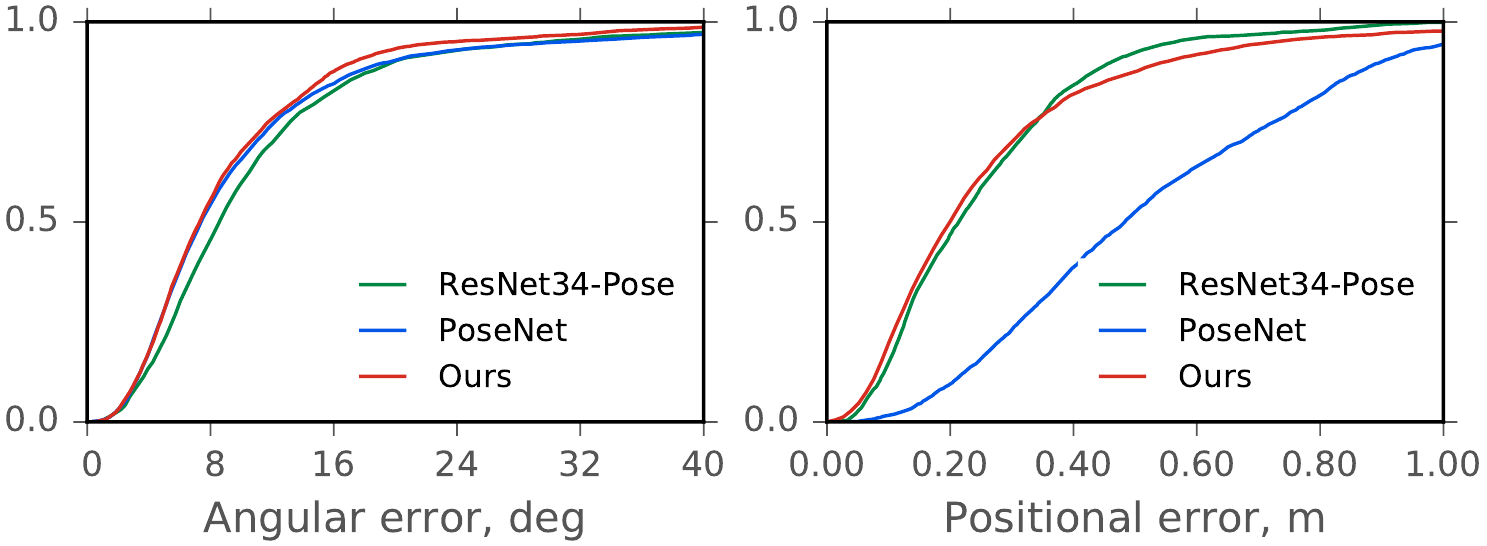}
 		\caption{Office}
 	\end{subfigure}
 	~
 	\begin{subfigure}[t]{.5\textwidth}
 		\centering
 		\includegraphics[width=\textwidth]{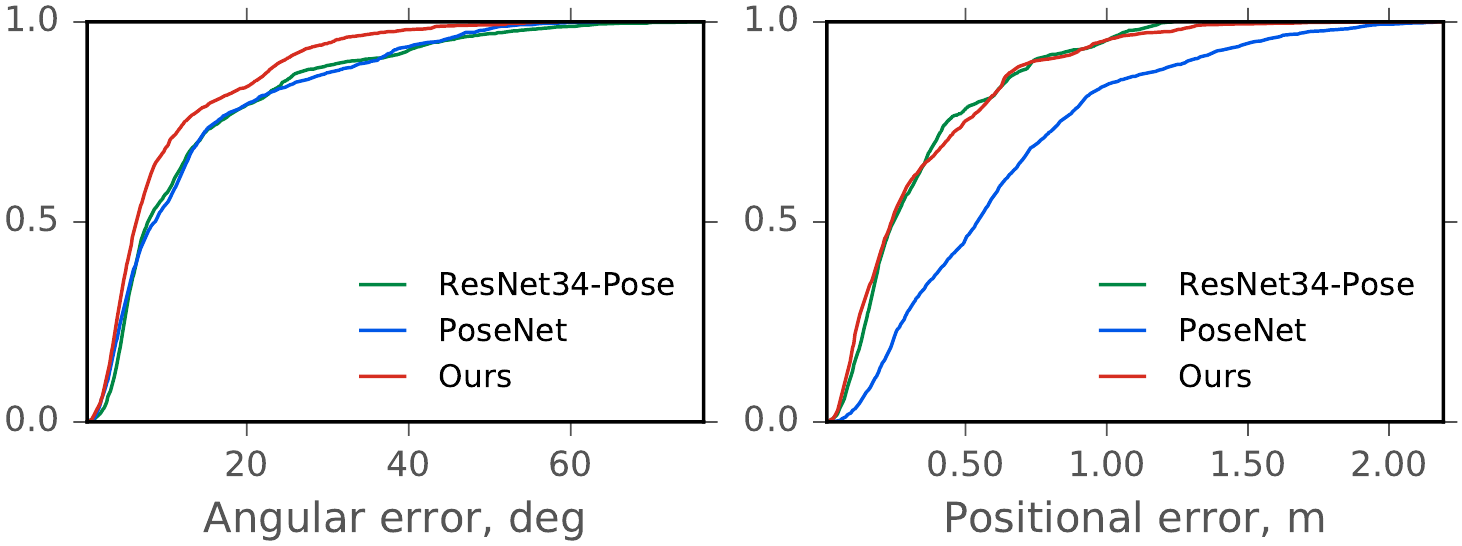}
 		\caption{Pumpkin}
 	\end{subfigure}%
 	~
 	\begin{subfigure}[t]{.5\textwidth}
 		\centering
 		\includegraphics[width=\textwidth]{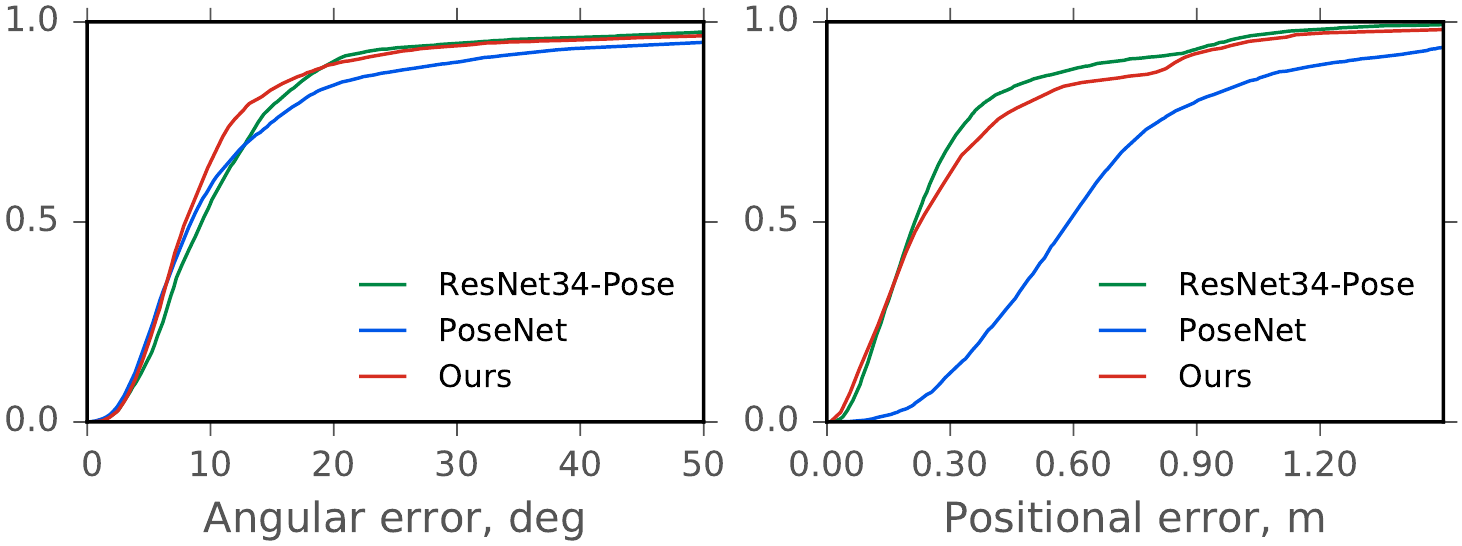}
 		\caption{Red Kitchen}
 	\end{subfigure}
 	~
 	\begin{subfigure}[t]{.5\textwidth}
 		\centering
 		\includegraphics[width=\textwidth]{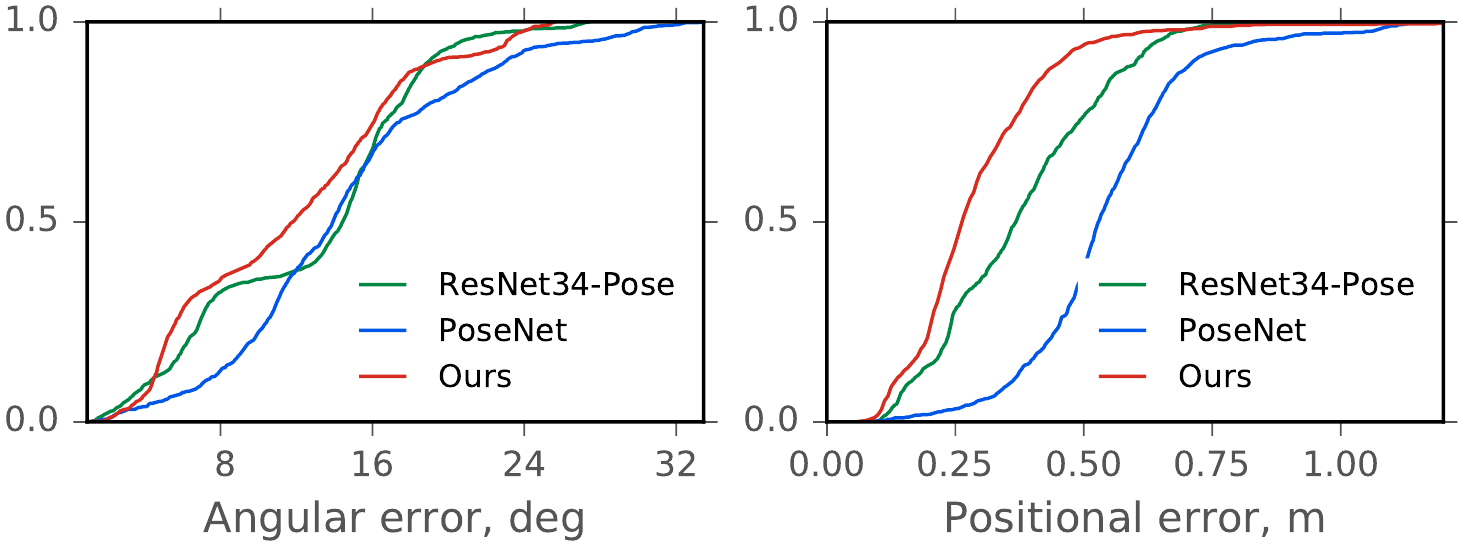}
 		\caption{Stairs}
 	\end{subfigure}
 \caption{Localization performance of the proposed approach, the baseline model (ResNet34-Pose), and PoseNet presented as normalized cumulative error histograms for all scenes of 7Scenes dataset. It should be noted that the baseline is trained in a scene-specific manner unlike ours. That is, our approach uses the same network for all 7 scenes whereas PoseNet and ResNet34-Pose have a separate network for each scene.
 }
 \end{figure*}
\end{document}